# Towards automatic identification of linguistic politeness in Hindi texts

**Ritesh Kumar**

Department of Linguistics
Dr. Bhim Rao Ambedkar University, Agra, India
riteshkrjnu@gmail.com

**Abstract**

In this paper I present a classifier for automatic identification of linguistic politeness in Hindi texts. I have used the manually annotated corpus of over 25,000 blog comments to train an SVM. Making use of the discursive and interactional approaches to politeness the paper gives an exposition of the normative, conventionalised politeness structures of Hindi. It is seen that using these manually recognised structures as features in training the SVM significantly improves the performance of the classifier on the test set. The trained system gives a significantly high accuracy of over 77% which is within 2% of human accuracy.

## 1. Introduction

Politeness is one of the most important components of human communication, which almost single-handedly decides whether the communication continues or it breaks mid-way. It is like the magnetic force which binds the interlocutors together. In the past few decades, politeness studies have taken a centre stage in the study of pragmatics with the publication of three seminal works on politeness – Lakoff (Lakoff, 1973), Brown & Levinson (1978, 1987), and Leech (Leech, 1983, 2007). However these theories are attacked on several counts by the discursive and interactional theorists who argue for a non-static, discursive approach to politeness studies.

In this paper I give a brief overview of the theoretical approaches of politeness followed by a brief analysis of polite structures in Hindi. I use this theoretical analysis for the construction of a classifier which could automatically recognise polite structures in Hindi.

## 2. Theories of politeness

Both Leech and Lakoff presents politeness in terms of Gricean maxims while B & L uses Goffman's concept of face to explain politeness across different cultures. Besides these, Leech also proposes a very significant distinction between two approaches to politeness studies- Absolute Politeness (renamed 'Semantic Politeness' in Leech (2007)) and Relative Politeness (renamed 'Pragmatic Politeness' in Leech (2007))

While semantic politeness refers to the approach which study the conventionalised, normative forms of politeness, pragmatic politeness refers to the non-normative, novel usage of politeness.

B & L have presented what could be called the most influential model of politeness. They define politeness as the mitigation of face-threatening acts (FTA). There are five superstragegies which could be employed by the speakers in order to do this – a) Do the FTA, without redressive action, baldly;  b) Positive politeness – It again contains fifteen strategies like pay attention to what the hearer is saying, joke to put the hearer at ease, use in-group identity markers, avoid disagreement, give gifts, etc.; c) Negative politeness – It contains ten strategies like be conventionally indirect, give deference, apologise, etc.; d) Off record; e) Withhold the FTA

Despite an apparent difference in the mechanics of these two classical theories of politeness, all these theories share certain common assumptions and they stand on very similar grounds. It is these assumptions that later came under the criticism of what are called discursive approaches to politeness (Locher & Watts, 2005; Mills, 2003; R. Watts, Ide, & Ehlich, 2005; R. J. Watts, 2003). Some of the most attacked aspects of these theories include the universality assumption (various studies starting with (Ide, 1989; Matsumoto, 1989) have attacked these theories for assuming that politeness across different cultures and languages could be described using these theories, and any one theory in general), adopting a largely speaker-oriented theory (these theories are an exposition of what the speakers are supposed to do and how do they do what they do without any significant reference to what is the role of the hearer/addressee in such situations) and presenting a theory which is largely oblivious to the role of context and discourse in the perception and production of politeness and instead focusses on certain speech acts/strategies which are considered inherently polite (while these theories do mention that context plays a role in politeness there is no mechanism for integrating the role of context in these theories).

While addressing these issues and even more, the discursive approaches present a theory of what they call first-order politeness (politeness as is perceived by the speakers of the language (R. J. Watts, 2003)) which is not at all universal and which seeks to describe politeness in the way it is perceived by the speakers of the language itself and not by using any scientific theories (which is called second-order politeness). This approach posits that politeness is a matter of relational work (Locher & Watts, 2005), thereby implying that it is not just a face-saving act but also something which helps in maintaining the personal relationships intact and balanced (Watts, 2003). Moreover it is not something which is static; rather it is always a matter of discursive struggle and dispute and the evaluation of politeness/impoliteness is subject to continuous change and renegotiation as the discourse progresses.

While a strong discursive approach completely rejects any kind of inherent politeness residing in any linguistic expression and maintains that politeness/impoliteness resides in the discourse and the discursive struggle which forms part of this discourse, the weaker form of this approach (also called interactional approaches) presents a case for certain conventionalised forms of politeness which are used by the speakers across several discourses for the similar effect of politeness and impoliteness and so they appear to be inherently polite (Culpeper, 2010). (Bousfield & Grainger, 2010; Culpeper, 2010; Terkourafi, 2005).

The present work is based on the interactional approaches to politeness which incorporated within itself both the conventional as well as non-conventional aspects of politeness. Conventionalisation of certain linguistic structures for their association with certain politeness effects imply that a large part of politeness evaluation is based on the prior experience of the speakers. However at the same time it must be emphasised that politeness is not completely deterministic which is decided solely by the priors; rather it is also emergent in the discourse, influenced by the local discourse and contextual factors. Thus it is expected that any identification system based on prior experiences (ultimately this is what supervised machine learning tasks are all about) would be able to recognise only conventionalised politeness structures in the text.

Any lexico-syntactic structure could be conventionally mapped to one of the following politeness levels (inspired by Watts' distinction between politic and polite)

Neutral text: These texts are basically cannot characterised as being polite or impolite. The texts such as plain description or objective and scientific texts could be termed neutral.

Appropriate text: These are what Watts calls 'politic' texts. These are the unmarked constructions in any discourse and are actually caused by the most conventionalised of forms. However not using these forms would make the text marked and potentially open to impolite interpretation. In the case of Hindi the use of honorific form of pronoun with elders would be labelled 'appropriate' since its use is unmarked and goes unnoticed but its absence would be potentially impolite.

Polite text: These texts contain the less conventionalised and marked constructions (in that discourse) so that their absence would make the text unmarked. These are generally those constructions where the speakers are "more" polite than is conventionally required in such situations.

Impolite text: These are the texts which have the structures conventionally associated with impoliteness. The most common structure would be the use of slangs.

## 3. Related Work

Alexandrov, Blanco, Ponomareva, & Rosso, (2007) had presented a way to "transform the lexical-grammatical properties of a text and the subjective expert opinion" to certain "numerical estimations". They construct a model using three factors relevant for politeness evaluation: the first greeting, polite words and polite grammar forms. While greeting may have a numerical value of either 0 or 1, the other two may take any numerical value in between 0 and 1. Using these three factors, a series of polynomial models are constructed. The same approach is discussed more specifically in Alexandrov, Ponomareva, & Blanco (2008) where a regression model is developed and trained for politeness estimation.

The work most closely related to the present work is that of Danescu-Niculescu-Mizil, Sudhof, Jurafsky, Leskovec, & Potts (2013) for English. It describes a classifier for identifying politeness in requests. An SVM classifier was trained using a manually annotated corpus of Wikipedia requests and tested against a corpus of requests from StackExchange. The annotators were required to rank the texts on a politeness scale which is later discretised into four categories, as per the final score assigned to them. While the experiments described in the paper gives a pretty good accuracy, in the present paper we obtain similar results using a simpler method.

In the experiments discussed in the present paper, a more general dataset of blog comments has been used instead of a very limited dataset of only requests. Moreover the cut-off scores used in the experiments by Danescu-Niculescu-Mizil, et al. (2013) are largely arbitrary. However in the present paper the four discrete categories used for annotating the data are theoretically motivated and so more representative of the empirical facts. Furthermore the theoretical framework used by Danescu-Niculescu-Mizil, et al. (2013) has been proved to be inadequate in explaining the empirical facts related to politeness which could undermine further efforts to obtain better accuracy in the task.

## 4. Annotation of CMC corpus of Hindi (CO3H)

Computer-mediated communication (CMC) provides a very rich variety of linguistic data. The data for Hindi CMC corpus (CO3H) is collected automatically from six different sources - Hindi Blogs, Hindi Web Portals, Hindi emails, Hindi chats, Youtube comments and Public and private chat over the web (Kumar, 2012).

Out of these, the data for this study is taken from the comments on Hindi blogs. The data consists of all kinds of comments which include requests, offers, complements, criticism, etc. Furthermore comments from only those blogs are taken which have 2 or more comments such that some kind of communication among the commentators and in between the blogger and the commentators is expected. Thus it is a pretty heterogenous data as well as representative of general human communication. A total of around 26,000 comments are taken for the study.

This data is manually annotated with one of the four politeness categories as discussed in the previous section – neutral, appropriate, polite and impolite. The annotators were given one comment at a time and they were required to classify it into one of these classes. The annotators were given a guide as to what these four categories are supposed to mean. The guide was similar to the description of these four categories given in Section 2. However the intuitions regarding putting the text into one of these categories was completely that of the annotators.

The data was annotated by two annotators using a web-based annotation tool (hosted at http://sanskrit.jnu.ac.in/tagit/). In order to calculate the inter-annotator agreement in between the two annotators 150 texts were given to each of them independently before the actual annotation task was given to them. Both the annotators annotated these texts completely independently using only the instructions provided to them. They were not allowed to discuss the categories or the intuitions about the text with each other or with anybody else. Initially the agreement in between the two speakers was at a dismal 50%.

This low agreement was not so much because of disagreement on the judgments about politeness as because of a lack of clarity regarding the four categories in which the texts were to be classified. Consequently they were given fresh set of instructions and a more detailed manual containing the definition and a clearer description of what each category stands for. After this the annotators were given the same set of annotated texts and they were asked to make revisions in the annotations based on their revised understanding of the categories. This time the annotations by the two annotators agreed in around 80% of cases. This agreement is similar to what Danescu-Niculescu-Mizil et al (2013) had reported for a similar task of politeness annotation of English requests. So it was assumed that the annotators now had a fairly good and common understanding of these categories stand for and the remaining disagreement is on account of their different intuitions about the text and not because of a misunderstanding of the categories for annotation.

## 5. Conventionalised politeness structures in Hindi

Some of the structures which are conventionally associated with politeness in Hindi are discussed below

**1. Use of formulaic expressions** like ʃubʰkamnaɟẽ, bədʰai, ʃukrija, dʰənjəwɑd, abʰar, kripja, etc. in the text.

e.g. 1
IPA     səŋgitɑ ɟi     **dhənjwɑd** kɑrtʊn    ko
Gloss   Sangita HON    **thanks**   cartoon   ACC
IPA     bhi     ʃamil   kərne        ke lije
Gloss   also    include do           for
FT      Sangeeta ji thank you for including the cartoon also

**2. Use of the particle ɟi**
e.g. 2
IPA     əti         sũndər    rəcnɑ
Gloss   extremely   beautiful composition
IPA     **ɟi**      dhənjwɑd
Gloss   **HON**     thanks
FT      Extremely beautiful composition thanks

**3. Use of the subjunctive verb form**
Subjunctive form of the verb is formed by adding -ẽ suffix to the last element of the verbal complex (leaving the copula) in Hindi.
e.g.3
IPA     əgər     munɑsib  səmɟhe    to      muɟhe   bhi
Gloss   if       proper   think     then    i.ACC   also
IPA     əpne     səmɑɟ    mẽ        ʃamil   **kərẽ**
Gloss   own      society  in        include **do.SUBJ**
FT      If you think it to be proper then please include me also in your society.

e.g. 4
IPA     əgər        kəkchɑ mẽ   səb       widjɑrthi
Gloss   if          class  in   all       students
IPA     səməɟh      pɑ     rəhe hɛ        ɒr
Gloss   understand  ECV    VCONT AUX      and
IPA     ek          mɛ     hi   ənɑɽi     hũ
Gloss   one         i      PRT  stupid    AUX.1PER
IPA     to          phir   jɑne **dẽ**
Gloss   then        again  go   **ECV.SUBJ**
FT      If all the students in the class have understood and I am the only stupid one then please let it be

**4. Use of the conditionals**
These are the canonical conditional (if...then) sentences in Hindi.
e.g.5
IPA     ɑp          iski       ʃuruɑt   **jədi**   ɛse
Gloss   you.HON     this       beginning **if**    like this
IPA     kərte       **to**     ədhik     prəbhɑwi  hotɑ
Gloss   do          **then**   more      effective would
FT      If you would have begun it like this then it would have been more effective

e.g. 6
IPA     **jədi**    do         əsthɑnõ   pər       ɟo      ki
Gloss   **if**      two        places    on        REL     COMPL
IPA     bəhʊt       bɑrik      kəmijɑ    hɛ        dʊr     kər
Gloss   very        minute     drawbacks AUX       remove  do
IPA     lijɑ        jɑje       **to**    bəhut     əcchɑ   hogɑ
Gloss   ECV         ECV        **then**  very      good    be.FUT
FT      If the drawbacks at two places, which are very minute, are removed then it would be very good

**5. Use of the suggestion markers/deontics**
Deontics are indicated by the use of **cɑhije** in the verbal complex in Hindi
e.g. 7
IPA     prəjɑs   thik    hɛ      pər     mɑtrɑ õr   ləj
Gloss   attempt  good    AUX     but     metre and  beats
IPA     ko       səməɟhne kɑ     ɒr      prəjɑs     kərnɑ **cɑhije**
Gloss   ACC      understand of   more    effort     do    **DEO**
FT      It is a good attempt but you should make more efforts to understand the metre and beats.

e.g. 8
IPA     ənɑm         bhɑi     ko      zjɑdɑ
Gloss   anonymous    brother  ACC     excessive
IPA     krodh        nəhĩ     kərnɑ   **cɑhije**

| Gloss | anger | NEG | do | **DEO** |
|---|---|---|---|---|
| | Anonymous | brother | should | not carry |
| FT | excessive anger. | | | |

### 6. Use of the ability markers/epistemics
Epistemic are indicated by the use of **səknɑ** or one of its morphological forms in the verbal complex in Hindi.

e.g. 9
| IPA | swɑsthjə | se | səmbəndhit | kəbhi | bhi |
|---|---|---|---|---|---|
| Gloss | health | about | related | anytime | also |
| IPA | kisi | bhi | jɑnkɑri | ke lije | ɑp |
| Gloss | any | also | information | for | you.HON |
| IPA | phon | bhi | kər | **səkte** | hɛ |
| Gloss | phone | also | do | **EPI** | AUX |
| FT | For any kind of information related to the health anytime you could call give a call. | | | | |

e.g. 10
| IPA | həmɑre | nəje | eɡriketər | mẽ | ɑp |
|---|---|---|---|---|---|
| Gloss | our | new | aggregator | LOC | you.HON |
| IPA | ɜpne | blɒɡ | ko | nice | ke |
| Gloss | your | blog | ACC | below | GEN |
| IPA | lĩŋko | dwɑrɑ jɒɽ | | **səkte** | hɛ |
| Gloss | links | by connect | | **EPI** | AUX |
| FT | In our new aggregator you could connect your blog by the links below. | | | | |

### 7. Use of particles zəra / jəra and thoɽɑ
e..g 11
| IPA | **jəɾɑ** | je | bhi | pəɽhije | pɛse | ki |
|---|---|---|---|---|---|---|
| Gloss | **just** | this | also | read.HON | money | of |
| IPA | nəji | pəribhɑsɑ. | ɑp | ke | kəl | |
| Gloss | new | definition | you.HON | of | tomorrow | |
| IPA | ke | cərcɑ | ke lije | bəɽhijɑ | hɛ | |
| Gloss | of | discussion | for | good | AUX | |
| FT | Just read this also the new definition of money. It is good for your tomorrow's discussion. | | | | | |

e.g. 12
| IPA | bhɑwpurɳ | rəcnɑ… | pər |
|---|---|---|---|
| Gloss | emotional | composition | but |
| IPA | jəɾɑ | wərtəni | mẽ sudhɑr |
| Gloss | just | spelling | in correction |
| IPA | kər | lẽ… | ɒr phir |
| Gloss | do | ECV.SUBJ | and again |
| IPA | se | post | kər dẽ... |
| Gloss | INST | post | do ECV.SUBJ |
| FT | Emotional Composition.... but just make correction in spelling and post it again. | | |

### 8. Use of the honorific pronominals and verb form
The +honorific forms of the verbs are generally formed by adding -ie suffix to the TAM bearing element(s) of the verbal complex.

e.g. 13
| IPA | ʊmmid | hɛ | mɛ | bhi | kəbhi |
|---|---|---|---|---|---|
| Gloss | expected | is | i | also | sometime |
| IPA | ɛsɑ | likh | pɑʊŋɡɑ ɜɡər | koi | əspesəl |
| Gloss | like this | writeECV | if | any | special |
| IPA | tips | ho | to | zərur | bətɑieɡɑ |
| Gloss | tips | be | then | necessarily | tell.HON |
| FT | It is expected that I shall also be able to write like this some day… if there is some special tips then do tell me. | | | | |

e.g. 14
| IPA | mɛne | ek | ɒr | koʃiʃ | ki |
|---|---|---|---|---|---|
| Gloss | I.ERG | one | more | try | do |
| IPA | hɛ | ɜɡər | ɑp | ko | pəsə̃nd |
| Gloss | AUX | if | you.HON | ACC | like |
| IPA | ɑje | to | ʊtsɑh | ke lije | |
| Gloss | ECV | then | enthusiasm | for | |
| IPA | ɜpne | səndeʃ | jərʊr | dijije | |
| Gloss | own | message | necessarily | give.HON | |
| FT | I have tried once more if you like it then please do give your message for enthusiasm | | | | |

## 6. Automatic Identification of Politeness

Using this theoretical analysis of Hindi politeness with the machine learning techniques I have developed a system for automatic identification of politeness in Hindi texts such that they could be classified in one of the four classes – neutral, appropriate, polite and impolite.

### Training the Classifier

I have developed a Support Vector Machine (SVM) using a total of 25660 texts annotated by human annotators for these four classes. It is randomly divided into train, test and validation set in 70:10:20 ratio, thereby, using a total of 17962 texts for training. I compare three classifiers – two Bag of words model, one using unigram feature representation and the other using unigram and bigram feature representation and a third classifier which uses unigrams, bigrams and the manually identified linguistic structures (discussed in the previous section) as features. The unigram model serves as the baseline model in the experiments.

### Testing the Classifier

The classifier is validated using 5132 texts and finally tested using 2566 texts. Table 1 gives a comparison of the performance of different classifiers trained using different kinds of features. The performance is measured in terms of simple percentage scores. The performance of 'human annotators' simply refers to the inter-annotator agreement in between the two annotators. The results are as obtained on the test set.

| **Feature Set** | **Test** |
|---|---|
| Unigrams | 75.45% |
| Unigrams and Bigrams | 75.72% |
| Unigrams, Bigrams and Linguistic Structures | **77.55%** |
| Human Annotators | **79%** |

Table 1: Comparative Performance of Classifiers

## 7. Analysing the results

As we could see from the test results, using a combination of unigrams, bigrams and linguistic features gives an improvement of a little over 2% in comparison to the baseline classifier while using just bigrams and unigrams does not lead to very significant improvements in the performance. Thus using specific, conventionalised linguistic structures (recognised through linguistic analysis of Hindi corpus) as features, along with the more general unigram and bigram features gives a very significant improvement to the system.

However at the same time an analysis of the texts that have been wrongly classified shows that the structures used in those texts were either not present in the train set or they were not classified as the same category as in the test set. It is here that Brown and Levinson's theory fails to account for the situation and provide a solution. However the interactional approaches explains the situation perfectly and in fact the theory had predicted that all the politeness level of all the texts cannot be predicted based on the prior occurrence since all politeness is not conventionalised.

However it could be concluded that these conventionalised structures are indeed helpful in developing a politeness recognition tool for Hindi. Furthermore in its current state the performance of the classifier is within 2% of human performance (the inter-annotator agreement is taken as a proxy for human performance) which is very close to the performance of the current atate-of-the-art in politeness recognition (Danescu-Niculescu-Mizil et al (2013)).

## 8. The way ahead

The performance of the system could be further improved by recognising and using more conventionalised politeness structures in Hindi. Another grey area where the things could be improved is the use of a more robust system of actually recognising the presence of these linguistic structures in the text. At present fairly robust regular expressions are used to automatically recognise these structures in the text. While these expressions have a high recall, the precision is comparatively low. So more sophisticated morphological analysers and parsers are required for better recognition of these structures in the text. Finally it is necessary to automatically recognise the emergent characteristics of politeness which may not be always signalled by linguistic means for developing a more comprehensive politeness recognition system.